\documentclass[runningheads]{llncs}
\usepackage{cite}
\usepackage{amsmath,amssymb,amsfonts}
\usepackage{graphicx, enumitem, booktabs}
\usepackage{multirow}
\usepackage{textcomp}
\usepackage[ruled,vlined,linesnumbered,longend]{algorithm2e}
\usepackage[table]{xcolor}
\usepackage{tikz}
\newcommand{\tikzcircle}[2][red,fill=red]{\tikz[baseline=-0.5ex]\draw[#1,radius=#2] (0,0) circle ;}
\newcolumntype{L}[1]{>{\raggedright\let\newline\\\arraybackslash\hspace{0pt}}m{#1}}
\newcolumntype{C}[1]{>{\centering\let\newline\\\arraybackslash\hspace{0pt}}m{#1}}
\newcolumntype{R}[1]{>{\raggedleft\let\newline\\\arraybackslash\hspace{0pt}}m{#1}} 

\def\BibTeX{{\rm B\kern-.05em{\sc i\kern-.025em b}\kern-.08em
    T\kern-.1667em\lower.7ex\hbox{E}\kern-.125emX}}
\begin{document}

\title{COEBA: A Coevolutionary Bat Algorithm for Discrete Evolutionary Multitasking}
\titlerunning{Coevolutionary Bat Algorithm}

\author{Eneko Osaba\inst{1} \and Javier Del Ser\inst{1,2} \and Xin-She Yang\inst{3} \and \\ Andres Iglesias\inst{4,5} \and Akemi Galvez\inst{4,5}}

\authorrunning{E. Osaba et al.} 

\institute{TECNALIA, Basque Research and Technology \\ Alliance (BRTA), 48160 Derio, Spain
\and University of the Basque Country (UPV/EHU), 48013 Bilbao, Spain
\and Middlesex University London, The Burroughs, London NW4 4BT, United Kingdom
\and Universidad de Cantabria, 39005 Santander, Spain
\and Toho University, Funabashi, Japan\\
\email{eneko.osaba@tecnalia.com}}
\maketitle              

\begin{abstract}
Multitasking optimization is an emerging research field which has attracted lot of attention in the scientific community. The main purpose of this paradigm is how to solve multiple optimization problems or tasks simultaneously by conducting a single search process. The main catalyst for reaching this objective is to exploit possible synergies and complementarities among the tasks to be optimized, helping each other by virtue of the transfer of knowledge among them (thereby being referred to as Transfer Optimization). In this context, Evolutionary Multitasking addresses Transfer Optimization problems by resorting to concepts from Evolutionary Computation for simultaneous solving the tasks at hand. This work contributes to this trend by proposing a novel algorithmic scheme for dealing with multitasking environments. The proposed approach, coined as Coevolutionary Bat Algorithm, finds its inspiration in concepts from both co-evolutionary strategies and the metaheuristic Bat Algorithm. We compare the performance of our proposed method with that of its Multifactorial Evolutionary Algorithm counterpart over 15 different multitasking setups, composed by eight reference instances of the discrete Traveling Salesman Problem. The experimentation and results stemming therefrom support the main hypothesis of this study: the proposed Coevolutionary Bat Algorithm is a promising meta-heuristic for solving Evolutionary Multitasking scenarios.
\end{abstract}

\section{Introduction} \label{sec:intro}

By using as its inspiration concepts from Transfer Learning \cite{pan2009survey} and Multitask Learning \cite{caruana1997multitask}, Transfer Optimization is an incipient knowledge field, which has congregated an active scientific community in recent years \cite{ong2016evolutionary}. The principal idea behind this field is to exploit what has been learned through the optimization of one specific problem or task, when tackling of another related or unrelated optimization task. Due to its relative youth, Transfer Optimization has not been studied as deeply as other research areas. It has not been until these last years when the transferability of knowledge among tasks has become a priority among researchers from the Evolutionary Computation arena.

Within the Transfer Optimization paradigm, three separated categories can be identified: \textit{sequential transfer}, \textit{multitasking} and \textit{multiform optimization}. The first of these classes refers to those situations in which tasks are faced sequentially, assuming that for solving a new problem/instance, the knowledge collected when solving previous tasks is used as external information \cite{feng2015memes}. The second of these categories (\textit{Multitasking}) deals with different optimization tasks simultaneously by dynamically scrutinizing existing complementarities and synergies among them \cite{gupta2016genetic,wen2017parting}. Finally, \textit{multiform optimization} aims at solving a single problem by resorting to different alternative problem formulations, which are optimized simultaneously. In all these categories, there is a clear consensus in the community on the capital importance of the correlation among the tasks to be solved for positively capitalizing on the transfer of knowledge over the search \cite{gupta2015multifactorial}.

Among the three divisions pointed out above, \textit{multitasking} is the one that has arguably grasped most attention by the community. The study presented in this manuscript is focused on this specific category. Specifically, we focus on multitasking optimization through the perspective of Evolutionary Multitasking (EM, \cite{ong2016towards}). In short, EM tackles the simultaneous optimization of several optimization tasks by relying on concepts and methods from Evolutionary Computation \cite{back1997handbook,del2019bio}. In the last years, a particular flavor of EM grounded on the so-called Multifactorial Optimization strategy (MFO, \cite{gupta2015multifactorial}) has shown a superior efficiency when dealing with different environments involving several continuous, discrete, single-optimization and multi-objective optimization problems and tasks \cite{wang2019evolutionary,gong2019evolutionary,yu2019multifactorial,gupta2016multiobjective}. The majority of the literature related to this area is focused on a solver belonging to this flavor: the Multifactorial Evolutionary Algorithm (MFEA, \cite{gupta2015multifactorial}). Unfortunately, alternatives to MFEA still remain scarce to date. 

Bearing this in mind, the research work presented in what follows revolves on a novel EM meta-heuristic algorithm that adopts the Bat Algorithm (BA, \cite{yang2010new}) at its core. Specifically, we present a Coevolutionary Bat Algorithm (COEBA) for discrete evolutionary multitasking. Through this proposal, we take a step further over the state of the art by elaborating on a new research direction in two different directions. On the one hand, we contribute to the EM area by introducing a new efficient meta-heuristic scheme. It is important to point out here that, unlike most articles published so far around EM, COEBA does not find its inspirational source in the MFO paradigm. On the other hand, COEBA is the first attempt at using BA for Transfer Optimization.

It is also relevant to underscore here that the experimentation carried out in this paper considers a less studied discrete environment comprising different instances of the Traveling Salesman Problem (TSP, \cite{lawler1985traveling}). Concretely, we assess the performance of the proposed COEBA by comparing its performance to that obtained by MFEA. Our main purpose with this performance comparison is to elucidate that COEBA embodies a promising alternative to deal with EM scenarios. To this end, we have chosen 8 different TSP instances, giving rise to 15 multi-tasking environments with varying degrees of phenotypical relationship.

The rest of the paper is organized as follows. Section \ref{sec:back} introduces the background related to both Evolutionary Multitasking and the Bat Algorithm. Next, Section \ref{sec:COEBA} exposes in detail the main features of the proposed COEBA. The experimentation setup, analysis and discussion of the results are given in Section \ref{sec:exp}. The study ends in Section \ref{sec:conc} with conclusions and future research directions.

\section{Background} \label{sec:back}

This section is dedicated to providing a brief background on Evolutionary Multitasking (Section \ref{sec:back_EM}) and the Bat Algorithm (Section \ref{sec:back_BA}).

\subsection{Evolutionary Multitasking} \label{sec:back_EM}

In recent years, EM has arisen as a promising paradigm for facing simultaneous optimization tasks. There are two main features that motivated the first formulation of EM. The first one is the parallelism inherent to the population of individuals, which eases the management of diverse concurrent optimization tasks faced simultaneously. Thanks to this feature, latent synergies between tasks can be automatically harnessed during the solving process \cite{ong2016evolutionary}. The second feature is the continuous transfer of genetic material between the individuals, which allows all tasks to benefit from each other, even for those that are not strongly correlated with the rest of the pool \cite{gupta2015multifactorial}.

It is widely accepted that the concept of EM was only materialized through the vision of the MFO until late 2017 \cite{da2017evolutionary}. Today, this nascent research stream is receiving interesting contributions in terms of new algorithmic schemes, such as the Coevolutionary Multitasking scheme proposed in \cite{cheng2017coevolutionary}, or the multitasking multi-swarm optimization described in \cite{song2019multitasking}. Additional alternatives to MFEA have also been proposed, such as the multifactorial brain storm optimization algorithm presented in \cite{zheng2016multifactorial}, the Multifactorial Differential Evolution in \cite{feng2017empirical} or the hybrid particle swarm optimization-firefly algorithm introduced in  \cite{xiao2019multifactorial}. Despite these recently proposed methods, MFO and its related MFEA have monopolized the research activity around this field since its inception. In fact, the authors of MFEA have recently introduced an adaptive variant of MFEA, coined as MFEA-II, thereby eliciting the momentum played by this algorithm in the field \cite{bali2019multifactorial}.

Going into mathematical details, we can formulate EM as an environment in which $K$ tasks or problems should be optimized in a simultaneous fashion. This environment is characterized by the existence of as many search spaces as tasks. Thus, for the $k$-th task, its objective function $T_k$ is characterized as $f_k : \Omega_k \rightarrow \mathbb{R}$, where $\Omega_k$ represents the search space of $T_k$. Let us assume that all tasks are minimization problems, so that the main objective of EM is to find a set of solutions $\{\mathbf{x}_1,\dots,\mathbf{x}_K\}$ such that $\mathbf{x}_k = \arg \min_{\mathbf{x}\in\Omega_k} f_k(\mathbf{x})$. A crucial aspect to properly understand EM is that all individuals $\mathbf{x}_p$ in the population $P$ to be evolved belong to a unified search space $\Omega^U$ that relates to $\Omega_1$ to $\Omega_K$ by means of a encoding/decoding mapping functions $\xi_k: \Omega_k\mapsto \Omega^U$. Therefore, each individual $\mathbf{x}_p\in \Omega^U$ in $P$ can be decoded ($\xi_k^{-1}(\mathbf{x}_p)$) to represent a task-specific solution $\mathbf{x}_{p,k}$ for each of the $K$ tasks. Shifting our attention on MFO and MFEA, four different definitions are associated with each individual $\mathbf{x}_p$ of the population $P$: Factorial Cost, Factorial Rank, Scalar Fitness and Skill Factor. With the intention accommodating this work to the extension requisites, we refer interested readers to \cite{gupta2015multifactorial,bali2019multifactorial} for additional deeper details on how these definitions are exploited during the search over the unified space $\Omega^U$. 

Several significant works have been recently published around EM and MFO. In \cite{yuan2016evolutionary}, authors present an influential application of the MFEA to different discrete problems. This paper also introduces the discrete unified encoding, used as a reference in subsequent works. A related study is \cite{zhou2016evolutionary}, where MFEA was applied to the Vehicle Routing Problem. Gong et al. presented in \cite{gong2019evolutionary} and improved version of the MFEA, endowing the algorithm with a dynamic resource allocating strategy. An interesting discrete MFEA has been also developed in \cite{wang2019evolutionary} for the composition of semantic web services. Gupta et al. presented in \cite{gupta2016multiobjective} a multi-objective variant of MFEA, giving evidence of its efficiency on a real-world manufacturing process design problem. Finally, the work in \cite{yu2019multifactorial} follows a similar strategy by enhancing MFEA with the incorporation of opposition-based learning. Further theoretical studies on EM and MFEO can be found in \cite{li2018multipopulation,zhou2019towards}.

\subsection{Bat Algorithm} \label{sec:back_BA}

BA is a nature-inspired metaheuristic based on the echolocation system of bats. In the nature, bats emit ultrasonic pulses to the surrounding environment with navigation and hunting purposes. After the emission of these pulses, bats listen to the echoes, and based on them they can locate themselves and also identify and locate preys and obstacles. Besides that, each bat is able to find the most ``nutritious'' areas performing an individual search, or moving towards a ``nutritious'' location previously found by any other component of the swarm. It is important to mention that some rules have to be previously established with the aim of making an appropriate adaptation \cite{yang2010new}:  
\begin{enumerate}
	\item All bats use echolocation to detect the distance, and they are assumed to be able to distinguish between an obstacle and a prey.
	\item All bats fly randomly at speed $v_i$ and position $\mathbf{x}_i$, emitting pulses with a fixed frequency $f_{min}$, varying wavelength $\lambda$ and loudness $A_i$ to search for a prey. In this idealized rule, it is assumed that every bat can adjust in an automatic way the frequency (or wavelength) of the pulses, emitted at a rate $r\in[0,1]$. This automatic adjustment depends on the proximity of the targeted prey.
	\item In the real world, the bats' emissions loudness can vary in many different ways. Nevertheless, we assume that this loudness can vary from a large positive $A_0$ to a minimum constant value $A_{min}$.
\end{enumerate}

Since its proposal, BA has emerged as one of the most successful meta-heuristic solvers. It has been applied to a manifold of problems such as logistic \cite{osaba2019discrete}, industry \cite{lu2019bi}, or medicine \cite{ibrahim2019enhanced}. The literature behind BA is huge and diverse, as manifested by comprehensive surveys on practical applications of BA \cite{yang2013bat,fister2013brief}.

\section{Coevolutionary Bat Algorithm for Multitasking}\label{sec:COEBA}

Following concepts previously embraced by other alternatives in the literature \cite{cheng2017coevolutionary}, one of the main characteristics of the designed COEBA is its multi-population nature. By this we mean that COEBA is a method composed by a defined number of populations, or \emph{demes} \cite{luque2011parallel}, comprised by the same number of individuals. More specifically, the number of groups is equal to $K$, i.e. the number of tasks to be optimized. Additionally, each of the $K$ subpopulations concentrates on solving a specific task $T_k$. This means that bats corresponding to the $k$-th deme are only evaluated on task $T_k$.

As in MFEA, a unified representation $\Omega^U$ is used for encoding individuals. However, the most innovative aspect of COEBA is that each subpopulation has its own search space. This involves a slight size readjustment when different demes exchange individuals among them. We will hereafter use the TSP to show this size readjustment problem. Hence, we denote the size of each problem $T_k$ (i.e. the number of \emph{cities}) as $D_k$. Let us assume that individual $\mathbf{x}_i$ is encoded as a permutation of the integer set $\{1,2,\ldots, D_{k}\}$. In this way, when $\mathbf{x}_p^k\in\Omega_k$ is migrated to a subpopulation in which the size of task $T_k'$ to be solved is $D_k'<D_k$, only integers lower than $D_k$ are considered, thus reducing the phenotype of the individual. These integers maintain the same order as in \smash{$\mathbf{x}_p^k$}. The reverse procedure applies if $D_k'>D_k$. In that case, and considering that each time an individual \smash{$\mathbf{x}_p^k$} is transferred to a deme it replaces an alternative bat \smash{$\mathbf{x}^{k'}_p$}, all integers between $D_k$ and $D_k'$ are inserted in $\mathbf{x}^k_p$ in the same positions as in \smash{$\mathbf{x}^{k'}_p$}. This multiple search space strategy enhances the exploitation of the search over the demes, making the movement operators more effective.

With all these considerations in mind, Algorithm \ref{alg:COEBA} shows the pseudocode of the designed COEBA. As can be seen, in the initialization process a number $X$ of individuals are randomly generated. After initialization, each individual is evaluated over all $K$ tasks. Then, within an iterative process, each subpopulation is built by choosing the top $X/K$ individuals for the corresponding task (the same bat can be selected by different tasks). Once demes are composed, each one is evolved independently by following the concepts of the discrete version of the BA \cite{osaba2019discrete}. To be more concise, the distance between two different bats is measured by means of the Hamming Distance, namely, the number of non-corresponding elements in the sequence. Furthermore, the \textit{inclination} mechanism is also used \cite{osaba2016improved}. Thanks to this feature (lines 10-14 in Algorithm \ref{alg:COEBA}), the method intelligently selects the movement function suited for each bat at every iteration, depending on its specific situation regarding the leading bat of the swarm. As is shown in Algorithm \ref{alg:COEBA}, \textit{2-opt} and \textit{insertions} are used as movement functions.
\begin{algorithm}[h!]
	 \DontPrintSemicolon
		Randomly generate an initial population of $X$ bats\;
		\For{each bat $x_i$ in the population}{
				Initialize the pulse rate $r_i$, velocity $v_i$ and loudness $A_i$\; 
		}
		Evaluate each of the individual for all the $K$ optimization tasks\;
		Build the $K$ number of subpopulations\;
		
		\While{termination criterion not reached}{
		    \For{each population $k$}{
		        \For{each bat $x_i$ in the subpopulation}{
						Generate new solution\;
						\eIf{$v_i^t$$<$n/2}{
								$x_i \leftarrow 2-opt(x_i^{t-1},v_i^t$)\;
							}{
								$x_i \leftarrow insertion(x_i^{t-1},v_i^t$)\;
							}
						\If{rand$>$$r_i$}{
							Select one solution among the best ones\;
							Generate a new bat by selecting the best neighbor of the chosen bat\;
						}
						\If{rand$<$$A_i$ and $f(x_i)$$<$$f(x_*)$}{
						    Accept the new solution\;
							Increase $r_i$ and reduce $A_i$\;
						}
				}
			    \If{iteration is multiple of \textrm{migr}}{
			        Two random individuals are migrated from $k$ to another randomly selected subpopulation\;
		        }
		    }
		}
		Return the best individual in $P$ for each task $T_k$\;
   \caption{Pseudocode of the proposed COEBA}
	 \label{alg:COEBA}
\end{algorithm}

Moreover, every $migr$ iterations, each group transfers two individuals to a randomly selected population. These two bats are selected by following this criterion: the first one is selected uniformly at random among the best $10$ individuals of the population, while the second one is drawn from the complete subpopulation. These two individuals substitute two randomly chosen bats, not considering the 10 best ones of the deme where the replacement is done. Finally, COEBA finishes its execution after $I$ iterations, returning as its solution the best bat of each subpopulation. Any other stopping criterion can be adopted with no further consequences to the design of the algorithm. 

\section{Experimentation and Results}\label{sec:exp}

To assess the performance of the designed COEBA solver, an extensive experimentation has been carried out, which is described in depth in this section. As such, Subsection \ref{sec:exp_TSP} elaborates on the group of TSP instances used in the experiments, whereas Subsection \ref{sec:exp_setup} details the experimentation setup. Finally, the obtained results are analyzed and critically examined in Subsection \ref{sec:exp_res}.

\subsection{Benchmark Problems}\label{sec:exp_TSP}

As introduced in Section \ref{sec:intro}, the experiments performed in this work consider the TSP as their benchmark problem to be optimized simultaneously. Readers interested on the formulation and theoretical aspects of this classical problem are referred to \cite{bellmore1968traveling} or \cite{miller1960integer}. Arguably, TSP has become one of the most often used problems for performance analysis of discrete optimization algorithms. A plethora of meta-heuristic solvers have been applied to the TSP, or to any of its variants, from traditional techniques such as the Genetic Algorithm \cite{grefenstette1985genetic}, Ant Colony Optimization \cite{dorigo1997ant} or Tabu Search \cite{gendreau1998tabu}, to modern discrete solvers such as Firefly Algorithm \cite{kumbharana2013solving}, Cuckoo Search \cite{ouaarab2014discrete}, or the Water Cycle Algorithm \cite{osaba2018discrete}. Before proceeding further, it is important to bear in mind that the goal of the experiments is not to reach the optimal solution of the TSP instances under consideration, but to statistically compare the performance of both MFEA and COEBA when using the same instances and evaluation conditions.

This being said, the performance of COEBA and its counterpart MFEA has been measured over 8 TSP instances, which are combined to yield 15 different test scenarios. All instances have been retrieved from the TSPLIB repository \cite{TSPLib}. Specifically, the first 8 instances of the Padberg/Rinaldi benchmark have been employed: \texttt{pr76}, \texttt{pr107}, \texttt{pr124}, \texttt{pr136}, \texttt{pr144}, \texttt{pr152}, \texttt{pr226}, and \texttt{pr264}.

\subsection{Experimental Setup}\label{sec:exp_setup}

For the sake of fairness in the comparisons, similar parameters and operators have been used for both MFEA and COEBA. This way, we can objectively conclude which solver reaches better outcomes using similar evaluation conditions. To ensure the replicability of this study, parameters employed for the implemented algorithms are depicted in Table \ref{tab:Parametrization}. For this parameter setting, not only works focused on MFEA and BA have been considered \cite{yuan2016evolutionary,gupta2015multifactorial,yang2013bat}, but also good practices reported in the community for tackling routing problems \cite{osaba2018good}. In addition, all bats are initialized uniformly at random. As the termination criterion, each algorithm is stopped after $I=500\cdot 10^3$ objective function evaluations.
\begin{table}[h!]
	\centering
	\vspace{-0.3cm}
	    \renewcommand{\arraystretch}{1}
	\resizebox{1.0\columnwidth}{!}{
		\begin{tabular}{lC{4.5cm}C{0.5cm}lC{4.5cm}}
			\toprule 
			\multicolumn{2}{c}{MFEA} & & \multicolumn{2}{c}{COEBA} \\
			\cmidrule{1-2} \cmidrule{4-5}
			Parameter & Value & & Parameter & Value\\
			\cmidrule{1-2} \cmidrule{4-5}
			Population size & 200 & & Population size & 200 \\ 
			Crossover Function & Order Crossover \cite{davis1985job} & & Short movement function & 2-opt \cite{lin1965computer}\\ 
			Mutation Function & 2-opt \cite{lin1965computer} & & Short movement function & Insertion \cite{osaba2016improved}\\
			Crossover Prob. & 0.9 & & Initial $A_i^0$ & Random number in [0.8,1.0]  \\
			Mutation Prob. & 0.1 & & Initial $r_i^0$ & Random number in [0.0,0.4]  \\
		    $migr$ & 100 & & $migr$ & 100 \\
		    & & & $\alpha$ \& $\gamma$ & 0.98 \\ \bottomrule
		\end{tabular}
	}
	\vspace{2mm}
	\caption{Parameter values set for MFEA and COEBA.}
	\vspace{-5mm}
	\label{tab:Parametrization}
\end{table}

As mentioned before, 15 different test scenarios have been built for the experimentation. Each of these multitasking configurations implies that both COEBA and MFEA should face the resolution of all the tasks assigned to that scenario simultaneously. Among these test cases, 10 of them are composed by 4 TSP instances, 4 scenarios are comprised by 6 TSP instances, and the last one includes all the 8 instances. Table \ref{tab:testCases} summarizes all the considered configurations. The main rationale for building these tests scenarios is twofold: i) to reach conclusions over a diverse and heterogeneous set of multitasking scenarios, involving each TSP instance in exactly the same number of cases, and ii) to exploit the possible genetic complementarities of the instances.
\begin{table}[ht!]
    \centering
        \renewcommand{\arraystretch}{1.1}
    \resizebox{0.95\columnwidth}{!}{
        \begin{tabular}{C{2.5cm}C{1cm}C{1cm}C{1cm}C{1cm}C{1cm}C{1cm}C{1cm}C{1cm}}
        \toprule
            Test Case & \texttt{pr76} & \texttt{pr107} & \texttt{pr124} & \texttt{pr136} & \texttt{pr144} & \texttt{pr152} & \texttt{pr226} & \texttt{pr264}\\ 
            \midrule
            \emph{Test\_Case\_4\_1} & $\times$ & $\times$ & $\times$ & $\times$ &   &   &   &   \\ 
            \emph{Test\_Case\_4\_2} &   &   &   &   & $\times$ & $\times$ & $\times$ & $\times$ \\
            \emph{Test\_Case\_4\_3} & $\times$ & $\times$ &   &   &   &   & $\times$ & $\times$ \\
            \emph{Test\_Case\_4\_4} &   &   & $\times$ & $\times$ & $\times$ & $\times$ &   &   \\
            \emph{Test\_Case\_4\_5} & $\times$ &   & $\times$ & $\times$ &   &   & $\times$ &   \\
            \emph{Test\_Case\_4\_6} &   & $\times$ &   &   & $\times$ & $\times$ &   & $\times$ \\ 
            \emph{Test\_Case\_4\_7} & $\times$ & $\times$ &   & $\times$ &   & $\times$ &   &   \\
            \emph{Test\_Case\_4\_8} &   &   & $\times$ &   & $\times$ &   & $\times$ & $\times$ \\ 
            \emph{Test\_Case\_4\_9} & $\times$ &   &   & $\times$ & $\times$ &   & $\times$ &   \\
            \emph{Test\_Case\_4\_10} &  & $\times$ & $\times$ &   &   & $\times$ &   & $\times$ \\
            \emph{Test\_Case\_6\_1} & $\times$ & $\times$ & $\times$ & $\times$ & $\times$ & $\times$ &   &   \\ 
            \emph{Test\_Case\_6\_2} &   &   & $\times$ & $\times$ & $\times$ & $\times$ & $\times$ & $\times$ \\ 
            \emph{Test\_Case\_6\_3} & $\times$ & $\times$ &   &   & $\times$ & $\times$ & $\times$ & $\times$ \\ 
            \emph{Test\_Case\_6\_4} & $\times$ & $\times$ &   & $\times$ & $\times$ &   & $\times$ & $\times$ \\ 
            \emph{Test\_Case\_8} & $\times$ & $\times$ & $\times$ & $\times$ & $\times$ & $\times$ & $\times$ & $\times$ \\ 
            \bottomrule
        \end{tabular}
    }
    \vspace{2mm}
    \caption{Summary of the 15 tests cases built for the experimentation.}
    	\vspace{-8mm}
    \label{tab:testCases}
\end{table}

Finally, all tests have been carried out on an Intel Xeon E5 – 2650 v3 computer, with 2.30 GHz and a RAM of 32 GB. Moreover, each test case has been run 20 times to account for the statistical significance of performance gaps encountered during the experimentation. 

\subsection{Results and Discussion}\label{sec:exp_res}

Table \ref{tab:testCasesSummary} depicts the comparisons in the results reached by COEBA and MFEA. Due to lack of space, we omit all the average outcomes for each test case. Instead, we show graphically the comparison using two colored circles. A green circle ($\tikzcircle[fill=green]{3pt}$) implies that COEBA has performed better than MFEA in terms of fitness average. On the contrary, a red circle ($\tikzcircle[fill=red]{3pt}$) means that MFEA has achieved better results on average. Using \emph{Test\_Case\_4\_3} as an example, and considering Table \ref{tab:testCases}, we observe that COEBA performs better in \texttt{pr76}, \texttt{pr226}, and \texttt{pr264}, while MFEA is better in \texttt{pr107}. Thus, analyzing the content of the Table \ref{tab:testCasesSummary}, we conclude that COEBA clearly outperforms MFEA over these EM scenarios, being superior to MFEA in all but 4 TSP instances evolved jointly. It is also crucial to highlight that COEBA obtains better outcomes in all the eight instances evolved jointly in \emph{Test\_Case\_8}.
\begin{table}[ht!]
    \centering
    \renewcommand{\arraystretch}{1.1}
    \resizebox{0.65\columnwidth}{!}{
        \begin{tabular}{C{2.5cm}C{5cm}}
        \toprule
            Test Case & COEBA vs. MFEA comparison\\ \midrule
            \emph{Test\_Case\_4\_1} & \tikzcircle[fill=green]{4pt}-\tikzcircle[fill=green]{4pt}-\tikzcircle[fill=green]{4pt}-\tikzcircle[fill=green]{4pt} \\ 
            \emph{Test\_Case\_4\_2} & \tikzcircle[fill=green]{4pt}-\tikzcircle[fill=green]{4pt}-\tikzcircle[fill=green]{4pt}-\tikzcircle[fill=green]{4pt} \\ 
            \emph{Test\_Case\_4\_3} & \tikzcircle[fill=green]{4pt}-\tikzcircle[fill=red]{4pt}-\tikzcircle[fill=green]{4pt}-\tikzcircle[fill=green]{4pt} \\ 
            \emph{Test\_Case\_4\_4} & \tikzcircle[fill=green]{4pt}-\tikzcircle[fill=green]{4pt}-\tikzcircle[fill=green]{4pt}-\tikzcircle[fill=green]{4pt}\\ 
            \emph{Test\_Case\_4\_5} & \tikzcircle[fill=green]{4pt}-\tikzcircle[fill=green]{4pt}-\tikzcircle[fill=green]{4pt}-\tikzcircle[fill=green]{4pt}\\ 
            \emph{Test\_Case\_4\_6} & \tikzcircle[fill=green]{4pt}-\tikzcircle[fill=green]{4pt}-\tikzcircle[fill=red]{4pt}-\tikzcircle[fill=green]{4pt}\\ 
            \emph{Test\_Case\_4\_7} & \tikzcircle[fill=green]{4pt}-\tikzcircle[fill=green]{4pt}-\tikzcircle[fill=green]{4pt}-\tikzcircle[fill=green]{4pt}\\ 
            \emph{Test\_Case\_4\_8} & \tikzcircle[fill=green]{4pt}-\tikzcircle[fill=green]{4pt}-\tikzcircle[fill=green]{4pt}-\tikzcircle[fill=red]{4pt}\\ 
            \emph{Test\_Case\_4\_9} & \tikzcircle[fill=green]{4pt}-\tikzcircle[fill=green]{4pt}-\tikzcircle[fill=green]{4pt}-\tikzcircle[fill=green]{4pt}\\ 
            \emph{Test\_Case\_4\_10} & \tikzcircle[fill=green]{4pt}-\tikzcircle[fill=green]{4pt}-\tikzcircle[fill=green]{4pt}-\tikzcircle[fill=green]{4pt}\\ 
            \emph{Test\_Case\_6\_1}& \tikzcircle[fill=green]{4pt}-\tikzcircle[fill=green]{4pt}-\tikzcircle[fill=red]{4pt}-\tikzcircle[fill=green]{4pt}-\tikzcircle[fill=green]{4pt}-\tikzcircle[fill=green]{4pt}\\ 
            \emph{Test\_Case\_6\_2}& \tikzcircle[fill=green]{4pt}-\tikzcircle[fill=green]{4pt}-\tikzcircle[fill=green]{4pt}-\tikzcircle[fill=green]{4pt}-\tikzcircle[fill=green]{4pt}-\tikzcircle[fill=green]{4pt}\\ 
            \emph{Test\_Case\_6\_3}& \tikzcircle[fill=green]{4pt}-\tikzcircle[fill=green]{4pt}-\tikzcircle[fill=green]{4pt}-\tikzcircle[fill=green]{4pt}-\tikzcircle[fill=green]{4pt}-\tikzcircle[fill=green]{4pt}\\ 
            \emph{Test\_Case\_6\_4}& \tikzcircle[fill=green]{4pt}-\tikzcircle[fill=green]{4pt}-\tikzcircle[fill=green]{4pt}-\tikzcircle[fill=green]{4pt}-\tikzcircle[fill=green]{4pt}-\tikzcircle[fill=green]{4pt}\\ 
            \emph{Test\_Case\_8} & \tikzcircle[fill=green]{4pt}-\tikzcircle[fill=green]{4pt}-\tikzcircle[fill=green]{4pt}-\tikzcircle[fill=green]{4pt}-\tikzcircle[fill=green]{4pt}-\tikzcircle[fill=green]{4pt}-\tikzcircle[fill=green]{4pt}-\tikzcircle[fill=green]{4pt}\\ \bottomrule
        \end{tabular}
    }
    \vspace{2mm}
    \caption{Comparison of the results for the 15 tests cases built for the experimentation. ($\tikzcircle[fill=green]{3pt}$) means COEBA outperforms MFEA. ($\tikzcircle[fill=red]{3pt}$) means MFEA performs better.}
    	\vspace{-5mm}
    \label{tab:testCasesSummary}
\end{table}

For extending the coverage and insights provided by this experimentation, we depict in Table \ref{tab:bestSolutions} the outcomes obtained by COEBA and MFEA for the 8 TSP instances that compose \textit{Test\_Case\_8}. We show the average, best and standard deviation of results for each instance. Furthermore, we also provide the best known optima reported for each TSP instance in the literature. These results confirm that the proposed COEBA is a promising meta-heuristic for solving EM environments, outperforming MFEA in terms of both average and best outcomes in this context. Even though it is not the goal of this work, it is also relevant to note that the difference between the optimal outcomes and the average results obtained by COEBA ranges between 0.4\% and 5.6\%, thereby showing that our proposal not only performs competitively for multitasking environments, but also gets close to optimality of the tasks under consideration.

In order to buttress our conclusions with the statistical significance of these identified gaps, the Wilcoxon Rank-Sum test has been applied, rendering the results depicted in the last row of Table \ref{tab:bestSolutions}. The confidence interval has been set to 95\%. We have compared the outcomes obtained for all the 8 TSP instances separately. Accordingly, the last row of Table \ref{tab:bestSolutions} represent the outcomes of these statistical tests. Specifically, a green circle ($\tikzcircle[fill=green]{3pt}$) means that COEBA outperforms MFEA with statistical significance. On the contrary, the red circle ($\tikzcircle[fill=red]{3pt}$) would have indicated the non-existence of evidences for ensuring the statistical significance of a gap between MFEA and COEBA. As can be seen in this table, Wilcoxon Rank-Sum test confirms that COEBA significantly outperforms MFEA in all the 8 instances embedded in this test scenario. The obtained average $z$-value is $-2.68$, with an average $p$-value equal to $0.00888$. Considering that the critical $z_c$ value is $-1.64$, and because $-2.68<-1.64$ and $0.00888<0.05$, these outcomes support the significance of the performance differences at $95$\% confidence level. Thus, the difference is significant at this confidence level, thereby concluding that the COEBA is statistically better than MFEA for this test scenario.
\begin{table}[h!]
    \centering
        \renewcommand{\arraystretch}{1.1}
    \resizebox{\columnwidth}{!}{
        \begin{tabular}{ccccccccc}
        \toprule
            Method & \texttt{pr76} & \texttt{pr107} & \texttt{pr124} & \texttt{pr136} & \texttt{pr144} & \texttt{pr152} & \texttt{pr226} & \texttt{pr264} \\ \midrule
            \multirow{3}{*}{COEBA} & \textbf{108602.4} & \textbf{44927.3} & \textbf{59380.8} & \textbf{99741.1} & \textbf{59045.5} & \textbf{74819.1} & \textbf{81425.7} & \textbf{51924.3}\\ 
            
            & 108234.0 & 44610.0 & 59087.0 & 99741.1 & 58771.0 & 74000.0 & 81048.0 & 51079.0\\ 
            
            & 402.54 & 242.27 & 226.89 & 534.30 & 244.37 & 420.50 & 248.55 & 458.87\\ \midrule
            
            \multirow{3}{*}{MFEA} & 113116.5 & 47110.5 & 62104.2 & 106729.3 & 62179.2 & 76117.3 & 84586.3 & 54031.7 \\
            
            & 111073.0 & 46052.0 & 61419.0 & 104998.0 & 60534.0 & 74294.0 & 82320.0 & 52728.0 \\
            
            & 2355.08 & 858.99 & 601.06 & 1461.53 & 1770.25 & 1756.13 & 6065.24 & 3489.31 \\ \midrule
            
            Optima & 108159.0 &  44303.0 & 59030.0 & 96772.0 & 58537.0 & 73682.0 & 80369.0 & 49135.0 \\
			\midrule
			Wilcoxon test & \tikzcircle[fill=green]{4pt} & \tikzcircle[fill=green]{4pt} & \tikzcircle[fill=green]{4pt} & \tikzcircle[fill=green]{4pt} & \tikzcircle[fill=green]{4pt} & \tikzcircle[fill=green]{4pt} & \tikzcircle[fill=green]{4pt} & \tikzcircle[fill=green]{4pt} \\ \bottomrule
            \end{tabular}
    }
    \vspace{2mm}
    \caption{Results (best/average/standard deviation of the fitness over 20 runs) obtained by COEBA and MFEA for the 8 instances in the $Test\_Case\_8$, and results of the Wilcoxon Rank-Sum test. Best results between COEBA and MFEA achieved over each TSP instances are highlighted in bold.}
    	\vspace{-10mm}
    \label{tab:bestSolutions}
\end{table}

\section{Conclusions and Future Work}\label{sec:conc}

This manuscript has elaborated on the design, implementation and validation of a novel approach for solving evolutionary multitasking environments, wherein tasks are optimization problems. For reaching this goal, we have introduced the Coevolutionary Bat Algorithm (COEBA), which finds its source of inspiration from the concepts of evolutionary co-evolution and the discrete adaptation of the Bat Algorithm. A subpopulation is devoted for the optimization of each problem, with a migration policy that allow exchanging genotype information and exploiting synergies among problems. For showcasing the application of the proposed multitasking approach, an experimental setup has been devised embracing instances of the Traveling Salesman Problem as benchmark problems to be jointly solved. We have compared the outcomes attained by COEBA with the ones furnished by Multifactorial Evolutionary Algorithm (MFEA) over 15 different multitasking test cases. The results validate our hypothesis: COEBA is a promising meta-heuristic for addressing multitasking scenarios.

Several research lines have been planned to gain insight beyond the findings reported in this study. In the short term, we will gauge the scalability of COEBA by analyzing its performance and computational efficiency when simultaneously solving test cases comprising TSP instances of larger dimensionality. We also plan to design additional search mechanisms (such as alternative migration strategies), all targeted at reinforcing the transfer of knowledge among related tasks (\emph{positive transfer}), and lowering the genotype exchange among those tasks that are not related to each other (correspondingly, \emph{negative transfer}). In the longer term, we will explore the application of COEBA to problems stemming from other research fields \cite{precup2019nature} with discrete optimization problems at their core, such as community detection in social networks.

\section*{Acknowledgments}

Eneko Osaba and Javier Del Ser would like to thank the Basque Government for its support through the EMAITEK and ELKARTEK programs. Javier Del Ser receives support from the Consolidated Research Group MATHMODE (IT1294-19) granted by the Department of Education of the Basque Government. Andres Iglesias and Akemi Galvez thank the Computer Science National Program of the Spanish Research Agency and European Funds, Project \#TIN2017-89275-R (AEI/FEDER, UE), and the PDE-GIR project of the European Union’s Horizon 2020 programme, Marie Sklodowska-Curie Actions grant agreement \#778035. 

\bibliographystyle{./bibliography/IEEEtran}
\bibliography{./bibliography/IEEEexample}

\end{document}